# Iterative Auto-Annotation for Scientific Named Entity Recognition Using BERT-Based Models

Kartik Gupta

**Abstract.** This paper presents an iterative approach to performing Scientific Named Entity Recognition (SciNER) using BERT-based models. We leverage transfer learning to fine-tune pre-trained models with a small but high-quality set of manually annotated data. The process is iteratively refined by using the fine-tuned model to auto-annotate a larger dataset, followed by additional rounds of fine-tuning. We evaluated two models, dslim/bert-large-NER and bert-large-cased, and found that bert-large-cased consistently outperformed the former. Our approach demonstrated significant improvements in prediction accuracy and F1 scores, especially for less common entity classes. Future work could include pre-training with unlabeled data, exploring more powerful encoders like RoBERTa, and expanding the scope of manual annotations. This methodology has broader applications in NLP tasks where access to labeled data is limited.

## 1 Introduction

Scientific Named Entity Recognition (SciNER), a fundamental task in natural language processing (NLP), is important for extracting structured information from scientific texts. The task involves identifying and categorizing specific entities such as method names, task names, dataset names, metric names and their values, hyperparameter names and their values within scientific publications in the NLP domain. Such a model can be used in applications such as automatic document parsing and document search and retrieval.

One of the critical challenges in developing a robust scientific entity recognition system is the availability of annotated data. Gathering large-scale annotated data for NLP tasks can be labor-intensive and time-consuming. In this report, we focus on NLP papers from recent NLP conferences such as ACL, EMNLP, and NAACL, as these papers provide the relevant information for our task. However, manually annotating thousands of papers from these conferences is not practical. To address this challenge, we first annotate a small set of data manually and use models trained on the small dataset to auto-annotate the remaining unlabeled data. Further, we repeat the auto-annotation process multiple times to improve the quality of auto-annotated data.

To train our models, we use Hugging Face Transformers library. We use the bert-large-cased variant of BERT family of encoder-only models to train our model on the trained dataset.

The paper is organized as follows: Section 2 describes the data collection process. Section 3 describes the annotation process. Section 4 describes the training of our model and auto-annotation of unlabeled data. Section 5 presents quantitative and qualitative analysis of models and Section 7 concludes the paper.

## 2 Data Collection

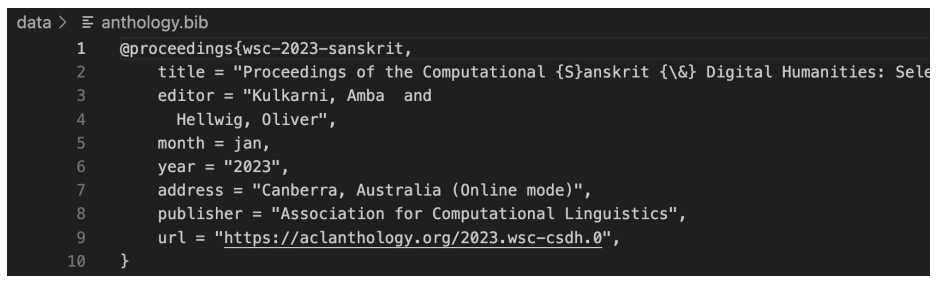

(a) anthology.bib

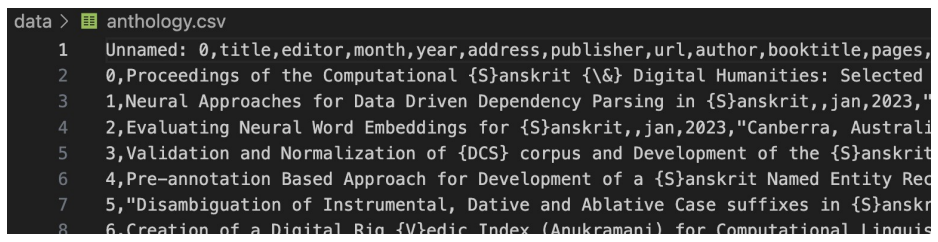

(b) anthology.csv

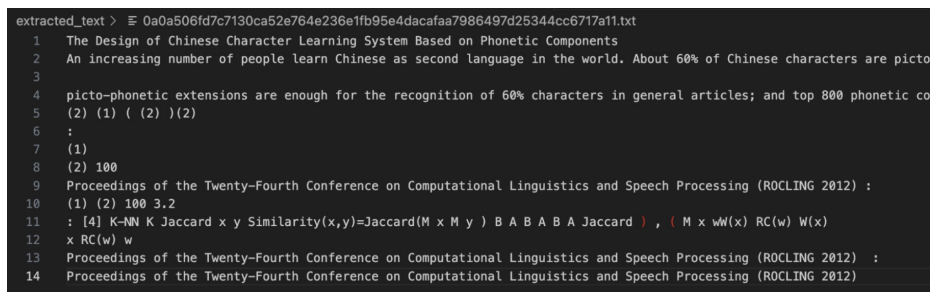

(c) extracted_text.txt

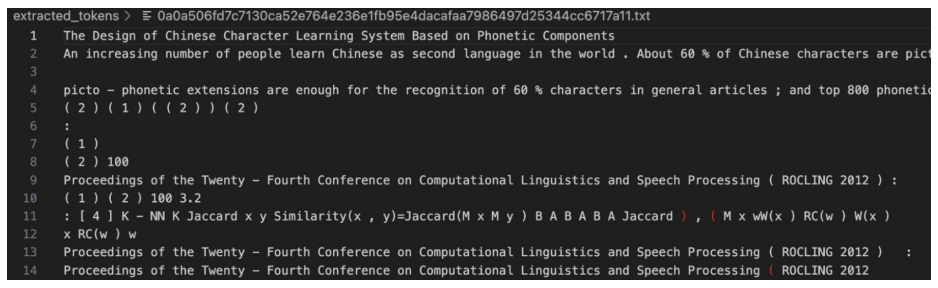

(d) extracted_tokens.txt

### 2.1 PDF Parsing

We chose ACL Anthology as our data source as it currently hosts 88,586 papers on the study of computational linguistics and natural language processing.

First, we use a Python script to parse the BibTeX filedownloaded from ACL Anthology website into a CSV file. Figure [1a]) shows a snapshot of the information contained in the file. The CSV file has columns that contain the title of the paper, year of publication, venue information (specifically NAACL, ACL, and EMNLP) for each paper and the URLs to download the PDFs. We hash the URL of the paper using SHA256 standard to uniquely identify a paper. A snapshot of the CSV file `anthology.csv` can be seen in the figure [1b].

After parsing the information about all the papers hosted on ACL Anthology, we create another script to automate the process of downloading PDF files from a list of URLs contained in the `anthology.csv` file. The hashes for URLs are used as file names for PDFs to ensure unique file names. We iterate through each row in CSV file to download and save PDFs. It is important to note that some of the PDFs failed to download, so we ended up downloading 87,587 (98.8%) of the PDFs.

After the PDFs are downloaded, we convert them into JSON files using SciPDF. We chose this library because it fits directly within our task for scientific publications. A snapshot of one such json file can be seen in the figure [1c].

### 2.2 Tokenizing

We use the spaCy tokenizer from the "en_core_web_lg" pre-trained model for the English language to tokenize our data. To speed up the tokenization process, we disable components in the spaCy pipeline such as "tok2vec," "tagger," "parser," "attribute_ruler,"

"lemmatizer," and "ner,". The tokenizer iterates through each json file in the input directory and tokenizes paper titles, paragraphs and other relevant sections. The tokenized text is stored in text files with tokenized text for one paragraph stored in each line. A snapshot of tokenized text can be seen in figure [1d].

To summarize, we begin with a bib file containing information about 88, 586 papers on ACL Anthology website and we extract tokenized data for all relevant paragraphs in 87, 587 of these papers.

# 3 Data Annotation

## 3.1 Methodology

We organized the 87,587 papers into three distinct categories based on the papers' publication venue, publication year and the requirement for manual annotation. The categories are manually annotated, automatically annotated, and unannotated. Manually annotated data contains 35 papers ACL, EMNLP, or NAACL conferences published in 2022 or 2023 that we manually annotated. All the remaining papers from these three conferences published in 2022 and 23 belong to the auto-annotated category. The rest of the papers that do not meet the criteria of publication venue and publication year belong to the unannotated data category.

We split the 35 manually annotated papers among the three members and we set aside 2 papers from each person for our test set, so that it adheres to variances in our annotation styles. In total we annotated about 1542 paragraphs manually using the Label Studio interface.

We also make use of close to 86,000 paragraphs from the auto-annotated category described above in the following way: We first train a NER model using the small labeled dataset and use the trained model to label these 86,000 paragraphs automatically. More details on auto-annotation can be found in the Model Training and Evaluation section [sec 4].

## 3.2 Annotation Interface

We use the Label Studio interface to annotate our data. The labels in the configuration are shown in Fig. 2.

## 3.3 Annotation Analysis

The Fig 3 shows the number of labels of each kind. Label 'O' is omitted since an overwhelming number of labels belong to that class 'O'.

# 4 Model Training and Evaluation

## 4.1 Model Details

In our proposed system for the Scientific NER task, we use transfer learning to finetune pre-trained models with manually annotated data. This approach shifts the primary challenge for human annotators away from requiring large volumes of annotated data and towards creating small but high-quality annotations. This strategy is suitable given the limited number of annotators available (three) and the time constraints (a few hours per week). We hypothesize that providing a relatively small set of high-quality and consistent manual annotations would steer the pre-existing knowledge to generalize to a broader test dataset.

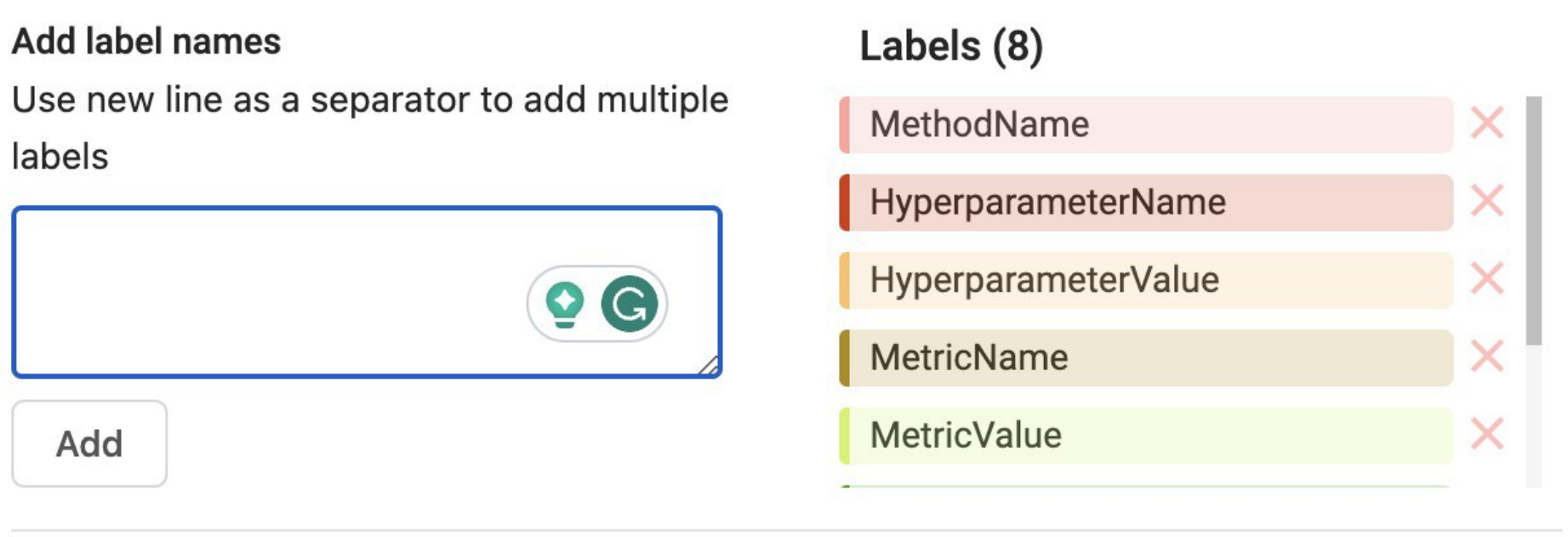

Fig. 2: Labeling Interface

In selecting a pre-trained model for this task, we consider a few important factors. The first factor is the amount of data the model is pre-trained on. We also consider two other factors: how easy it is to fine-tune the model to our specific manual annotation standard, and how well the NER task aligns with the pre-training objective. Due to these considerations, we decided not to explore decoder-only models like LLMs, as their generative nature leads to a different, prompting-based approach to fine-tuning that does not align naturally with the NER task. Instead, we focused on encoder-only models, evaluating the standard BERT [8] model along with its derivatives such as DeBERTa [9] and RoBERTa [10]. Based on the factors mentioned above, we selected the following pre-trained BERT variants available on HuggingFace: `dslim/bert-large-NER`, `bert-large-cased` and `roberta-large`.

The `bert-large-NER` is a specialized version of the BERT model, tailored for Named Entity Recognition (NER) tasks. It's trained to identify four entity categories: location (LOC), organizations (ORG), person (PER), and Miscellaneous (MISC). This model evolved from the `bert-large-cased` model by fine-tuning it on the CoNLL-2003 English dataset [12]. On the other hand, `bert-large-cased` is the original, case-sensitive BERT model, trained using Masked Language Modeling (MLM) and Next Sentence Prediction(NSP) objectives. For both models, a single fully connected layer is added at the end, acting as a classification head to transform BERT embeddings into probabilistic predictions for 15 classes in the SciNER task. We also tried experimenting with RoBERTa (using `roberta-large` on HuggingFace) in place of BERT, coupled with a 2-layer fully-connected classification head. However, this would require substantial alterations to the codebase, particularly the rule-based labeling scheme mentioned ahead, and the modifications didn't lead to significantly improved results. We, therefore, did not explore using RoBERTa encoder in our model.

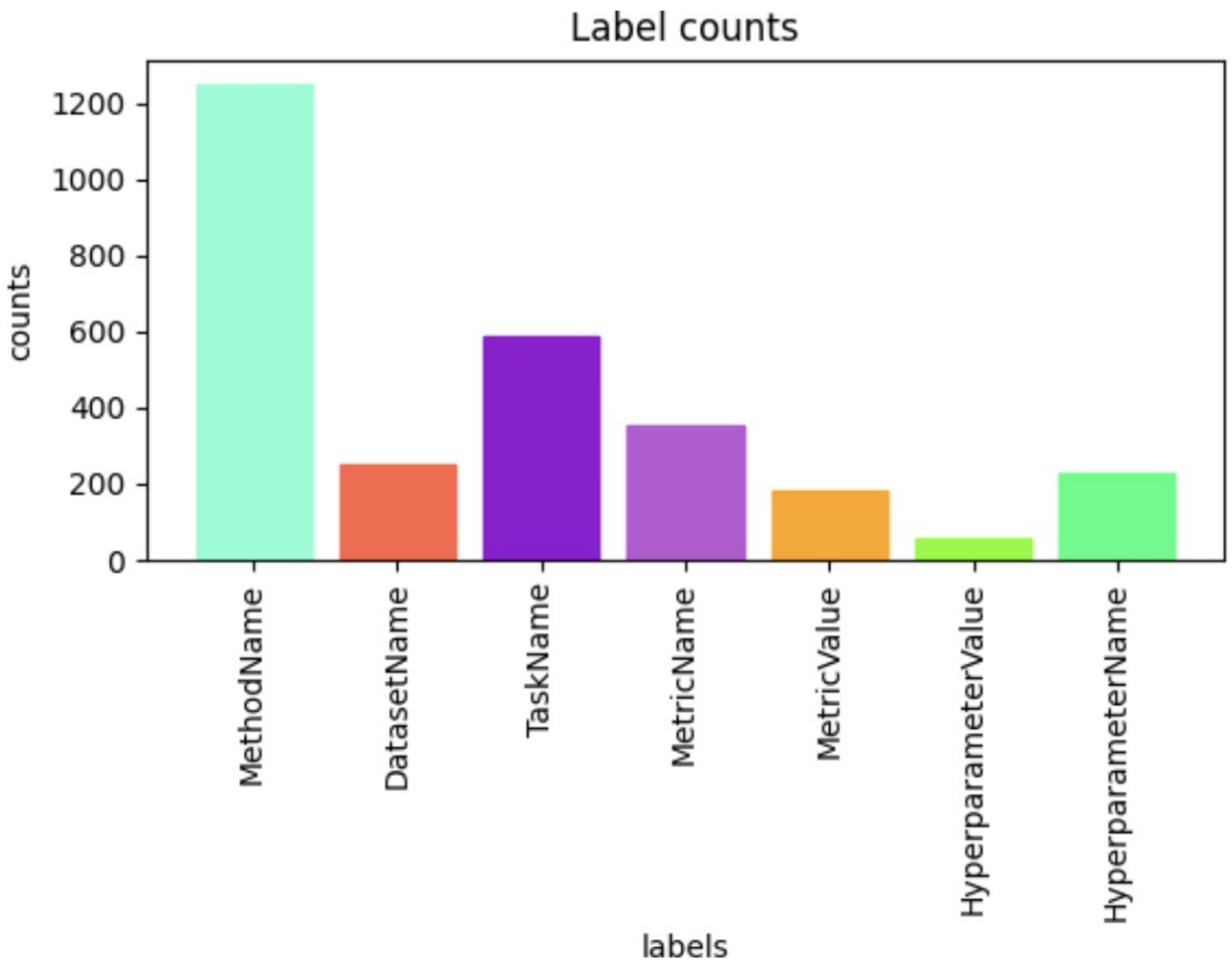

Fig. 3: Manually annotated label count

### 4.2 Training Process

As discussed previously, we use a model trained on a smaller manually annotated dataset to automatically annotate a larger dataset. This process of training on labeled data and using it to automatically annotate a larger dataset can be carried out in a loop to iteratively improve the quality of auto-annotated data. This process can be summarized in four steps as follows:

1. **Step-1**: Fine-tune the BERT-based model on manually annotated data.
2. **Step-2**: Use the fine-tuned model to generate automated annotations on the remaining corpus of scientific papers. Reject all annotations for which the model's confidence is less than a confidence threshold $\gamma$.
3. **Step-3**: Fine-tune the BERT model on a combination of manually and automatically annotated data.
4. **Step-4**: Repeat steps 1 through 3 multiple times.

In step 1, we provide the model with manually annotated paragraphs from three contributors. This annotation was carried out on 35 papers from the ACL anthology. A paragraph from a candidate research paper was chosen for annotation only if it was clear and understandable, adhering to the defined entity classes, and that the annotator could easily grasp the overall content of the paper. This method aimed to minimize labeling inconsistencies and allowed the model to tackle ambiguous cases using its built-in knowledge. The model is trained for 20 epochs with a learning rate of $1 \times 10^{-4}$ and a batch size of 8 on two NVIDIA RTX 3090 GPUs.

In step 2, we apply the model, now trained on manually annotated data, to annotate the labels for remaining ACL, EMNLP and NAACL papers from 2022-2023. We hypothesize that the model will reliably adhere to the manual annotation standard while working on unlabelled data. Since the predictions of the model are not always accurate, we use a confidence threshold to sift out uncertain annotations. Since BERT uses WordPiece tokenizer, we first aggregate class probabilities for each word by multiplying token-level probabilities. Once we have word-level class probabilities, we retain only those word-level predictions where the confidence ($p$) exceeds a confidence threshold of $\gamma = 0.98$. If not, the word is marked as ambiguous (‘*amb*').

$$p(\mathrm{c}|\mathrm{word}) = \prod_{i=1}^{N} p(\mathrm{c}|\mathrm{token}^{i}_{\mathrm{word}}) \quad (1)$$

where:
$N$ = total number of tokens in the word
$p(\mathrm{c}|\mathrm{token}^{i}_{\mathrm{word}})$ = probability that the token $t$ belongs to class c.

$$\mathrm{label}(\mathrm{word}) = \begin{cases} C & \text{if } \max_c p(\mathrm{c}|\mathrm{word}) \geq \gamma \\ `amb' & \text{otherwise} \end{cases} \quad (2)$$

where:
$C = \arg\max_c p(\mathrm{c}|\mathrm{word})$

In addition, we also use rule-based prediction to make our model predictions more accurate. For example, we know that a word with class ’0’ can only be followed by a word with classes ’O’ or ’B-<class name>’ but not ’I-<class name>’. Similarly, a word with class ’I-<class name 1>’ cannot be followed by a word with ’I-<class name 2>’. We use these rules inherent to SciNER task definition in our prediction process.

In step 3, the last step before testing the model’s performance, we fine-tune the model on a combined set of manual and automated annotations for 5 epochs with a learning rate of $1 \times 10^{-4}$ and a batch size of 8.

In step 4, we loop through steps 1-3 multiple times to improve our outcomes. For our final model, we perform this iteration 2 times. In 4.3 we show how this iterative auto-annotation and re-training of our model improves the performance on the test set.

### 4.3 Experiment Evaluation

We evaluate the two candidate models on a test set created from 6 papers as discussed in section 3. As part of the statistical significance bootstrap test , we randomly draw a subset of 50 samples from the evaluation set (without replacement) for a total of 12 times. For each draw, we calculate the corresponding Accuracy and F1 score (precision, recall) from both the models. For each metric, we present the mean value and the obtained range in Table 1. Based on these results, we carry subsequent experiments with the `bert-large-cased` model.

| Metrics | dslim/bert-large-NER | bert-large-cased |
|---|---|---|
| Mean Accuracy (%) | 95.1 | **96.4** |
| Accuracy Range (%) | 94.5 - 95.8 | **95.8 - 97.0** |
| Mean F1 | 0.576 | **0.741** |
| F1 Range (%) | 0.526 - 0.624 | **0.696 - 0.787** |

Table 1: Significance test for `bert-large-cased` and `dslim/bert-large-NER` over 12 slices of evaluation data

Table 2 below contains metrics computed on test set for three different `bert-large-cased` variants: model trained only on manually labeled data, model trained on combined manually and auto annotated data at the end of iteration 1 and model trained on combined manually and auto annotated data at the end of iteration 2. We can clearly see the improvement in metrics as we iteratively improve the quality of auto-annotated data.

| Iteration | Accuracy | F1 |
|---|---|---|
| Before iteration 1 | 94.2 | 0.559 |
| After iteration 1 | 97.7 | 0.856 |
| After iteration 2 | **99.7** | **0.960** |

Table 2: Metrics at different points in training process for `bert-large-cased`

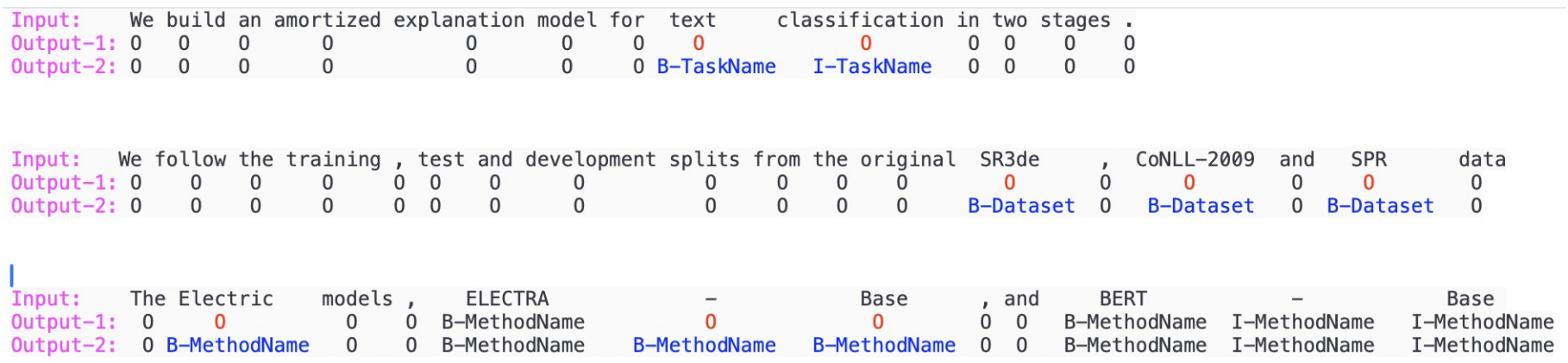

Fig. 4: Comparison of results before 1st iteration and after second iteration

## 5 Analysis

The improvements in overall metrics is reflected the most in predictions for less common classes. Fig 4 that compares results generated on examples from the test dataset proves this point. The figure shows three inputs and corresponding word-level predicted labels from models before first iteration and at the end of second iteration. We can clearly see the improvements in predictions of less common classes such as TaskName, DatasetName and MethodName. The incorrect and correct labels are highlighted in red and blue colors respectively.

## 6 Future Work

There are a few ways to improve the SciNER model further. The first way is by using the large quantity of data in "unlabeled" category to pre-train the model using MLM objective. This allows us to leverage tens of thousands of papers that haven't been annotated either manually or using an auto-annotation process. The second is by replacing BERT with other more powerful encoders such as RoBERTa. The third is by annotating a wider set of papers manually rather than restricting the manual annotation to just 35 papers.

## 7 Conclusion

In this work, we describe our approach to performing the SciNER task. Our iterative auto-annotation approach has applications beyond the NER task and has the potential to improve the performance of NLP tasks where getting access to labeled data is hard.

## References

1. Aho, Alfred V. and Ullman, Jeffrey D., *The Theory of Parsing, Translation and Compiling*, vol. 1, Prentice-Hall, Englewood Cliffs, NJ, 1972.
2. American Psychological Association, *Publications Manual*, American Psychological Association, Washington, DC, 1983.
3. Chandra, Ashok K., Kozen, Dexter C. and Stockmeyer, Larry J., "Alternation," *Journal of the Association for Computing Machinery*, vol. 28, no. 1, pp. 114–133, 1981. doi:10.1145/322234.322243.
4. Andrew, Galen and Gao, Jianfeng, "Scalable training of L1-regularized log-linear models," in *Proceedings of the 24th International Conference on Machine Learning*, pp. 33–40, 2007.
5. Gusfield, Dan, *Algorithms on Strings, Trees and Sequences*, Cambridge University Press, Cambridge, UK, 1997.
6. Rasooli, Mohammad Sadegh and Tetreault, Joel R., "Yara Parser: A Fast and Accurate Dependency Parser," *Computing Research Repository*, vol. arXiv:1503.06733, version 2, 2015. [Online]. Available: http://arxiv.org/abs/1503.06733.
7. Ando, Rie Kubota and Zhang, Tong, "A Framework for Learning Predictive Structures from Multiple Tasks and Unlabeled Data," *Journal of Machine Learning Research*, vol. 6, pp. 1817–1853, 2005. JMLR.org.
8. Devlin, Jacob, Chang, Ming-Wei, Lee, Kenton and Toutanova, Kristina, "BERT: Pre-training of Deep Bidirectional Transformers for Language Understanding," in *North American Chapter of the Association for Computational Linguistics*, 2019.
9. He, Pengcheng, Liu, Xiaodong, Gao, Jianfeng and Chen, Weizhu, "DeBERTa: Decoding-enhanced BERT with Disentangled Attention," *ArXiv*, vol. abs/2006.03654, 2020. [Online]. Available: https://api.semanticscholar.org/CorpusID:219531210.

10. Liu, Yinhan, Ott, Myle, Goyal, Naman, Du, Jingfei, Joshi, Mandar, Chen, Danqi, Levy, Omer, Lewis, Mike, Zettlemoyer, Luke and Stoyanov, Veselin, "RoBERTa: A Robustly Optimized BERT Pretraining Approach," *ArXiv*, vol. abs/1907.11692, 2019. [Online]. Available: https://api.semanticscholar.org/CorpusID:198953378.
11. Lan, Zhenzhong, Chen, Mingda, Goodman, Sebastian, Gimpel, Kevin, Sharma, Piyush and Soricut, Radu, "ALBERT: A Lite BERT for Self-supervised Learning of Language Representations," *ArXiv*, vol. abs/1909.11942, 2019. [Online]. Available: https://api.semanticscholar.org/CorpusID:202888986.
12. Tjong Kim Sang, Erik F. and De Meulder, Fien, "Introduction to the CoNLL-2003 Shared Task: Language-Independent Named Entity Recognition," in *Proceedings of the Seventh Conference on Natural Language Learning at HLT-NAACL 2003*, pp. 142–147, 2003. [Online]. Available: https://aclanthology.org/W03-0419.